%% file: main.tex
\documentclass[letterpaper, 10 pt, journal, twoside]{ieeetran}

\makeatletter

\usepackage{graphicx}
\usepackage{cancel}
\usepackage{hyperref}
\usepackage{bm}
\usepackage{balance}

\usepackage{amssymb}
\usepackage{mathtools}
\usepackage{amsmath}
\usepackage{cases}
\usepackage{cite}
\usepackage{multirow}
\usepackage{comment}
\usepackage{derivative}

\usepackage{amsthm} % For remark environment
\theoremstyle{definition}

\newtheorem{remark}{Remark}

\usepackage{xcolor}
\definecolor{lgray}{gray}{0.30}
\usepackage{eso-pic}
\usepackage{hyperref}
\AddToShipoutPictureBG*{%
  \AtPageUpperLeft{%
    \setlength{\unitlength}{1mm}%
    \put(-18.5,-9){\makebox(\paperwidth,0)[c]{\parbox{0.8\textwidth}{IEEE Robotics and Automation Letters, 2026, vol. 11, no. 1, pp. 1-8}}}
  }
}
\AddToShipoutPictureBG*{%
  \AtPageUpperLeft{%
    \setlength{\unitlength}{1mm}%
    \put(-9.5,-14){\makebox(\paperwidth,0)[c]{\parbox{0.9\textwidth}{DOI: \href{https://ieeexplore.ieee.org/document/11235953}{10.1109/LRA.2025.3630871}}}}
  }
}

\title{Feedback-MPPI: Fast Sampling-Based MPC via \\ Rollout Differentiation -- Adios low-level controllers}
\author{Tommaso Belvedere, Michael Ziegltrum, Giulio Turrisi, Valerio Modugno
\thanks{T. Belvedere is with CNRS, Univ Rennes, Inria, IRISA, Campus de Beaulieu, Rennes, France. E-mail: {\tt\scriptsize tommaso.belvedere@irisa.fr}}
\thanks{M. Ziegltrum and V. Modugno are with the Department of Computer Science, University College London, Gower Street, WC1E 6BT, London, UK. E-mail: {\tt\scriptsize \{michael.ziegltrum.24, v.modugno\}@ucl.ac.uk}}%
\thanks{G. Turrisi is with the Dynamic Legged Systems Laboratory, Istituto
Italiano di Tecnologia (IIT), Genova, Italy. E-mail: {\tt\scriptsize giulio.turrisi@iit.it}}%}
\thanks{Code and videos available at:
\url{https://feedback-mppi.github.io}}

}

\input{mathsym.tex}

\begin{document}

\maketitle

\begin{abstract}
Model Predictive Path Integral control is a powerful sampling-based approach suitable for complex robotic tasks due to its flexibility in handling nonlinear dynamics and non-convex costs. However, its applicability in real-time, high-frequency robotic control scenarios is limited by computational demands. This paper introduces Feedback-MPPI (F-MPPI), a novel framework that augments standard MPPI by computing local linear feedback gains derived from sensitivity analysis inspired by Riccati-based feedback used in gradient-based MPC. These gains allow for rapid closed-loop corrections around the current state without requiring full re-optimization at each timestep. We demonstrate the effectiveness of F-MPPI through simulations and real-world experiments on two robotic platforms: a quadrupedal robot performing dynamic locomotion on uneven terrain and a quadrotor executing aggressive maneuvers with onboard computation. Results illustrate that incorporating local feedback significantly improves control performance and stability, enabling robust, high-frequency operation suitable for complex robotic systems.
\end{abstract}

\section{Introduction}

\IEEEPARstart{M}{odel} Predictive Control (MPC) was initially adopted in the process industry, where slow system dynamics of chemical plants and refineries made it feasible to solve optimal control problems online without strict real-time constraints~\cite{Qin2003}. In recent years, advances in computational hardware have enabled the widespread use of MPC in robotics, where system dynamics are significantly faster than in traditional applications. To meet real-time requirements, these implementations often rely on simplified internal models, such as reduced-order or template dynamics~\cite{Katayama2023}.
Despite ongoing efforts to accelerate computation, the optimization time required to solve an MPC problem at each control step remains a significant bottleneck, making the application of MPC to fast-paced and highly nonlinear robotic systems an open and active research topic \cite{Wensing2024}. These limitations are particularly pronounced in tasks requiring high-frequency control, such as agile motion generation and dynamic physical interaction.

To address this computational bottleneck, several approaches have been developed. Methods like distributed optimization and transformer-based constraint handling offer improvements but require significant framework modifications or specialized hardware~\cite{Amatucci2024, Adabag2024, Nguyen2024}. Another strategy approximates MPC policies directly, for example, using neural networks, enabling near-instantaneous control after initial training~\cite{Hertneck2018}.

\begin{figure}[t!]
    \centering
    \includegraphics[width=\columnwidth]{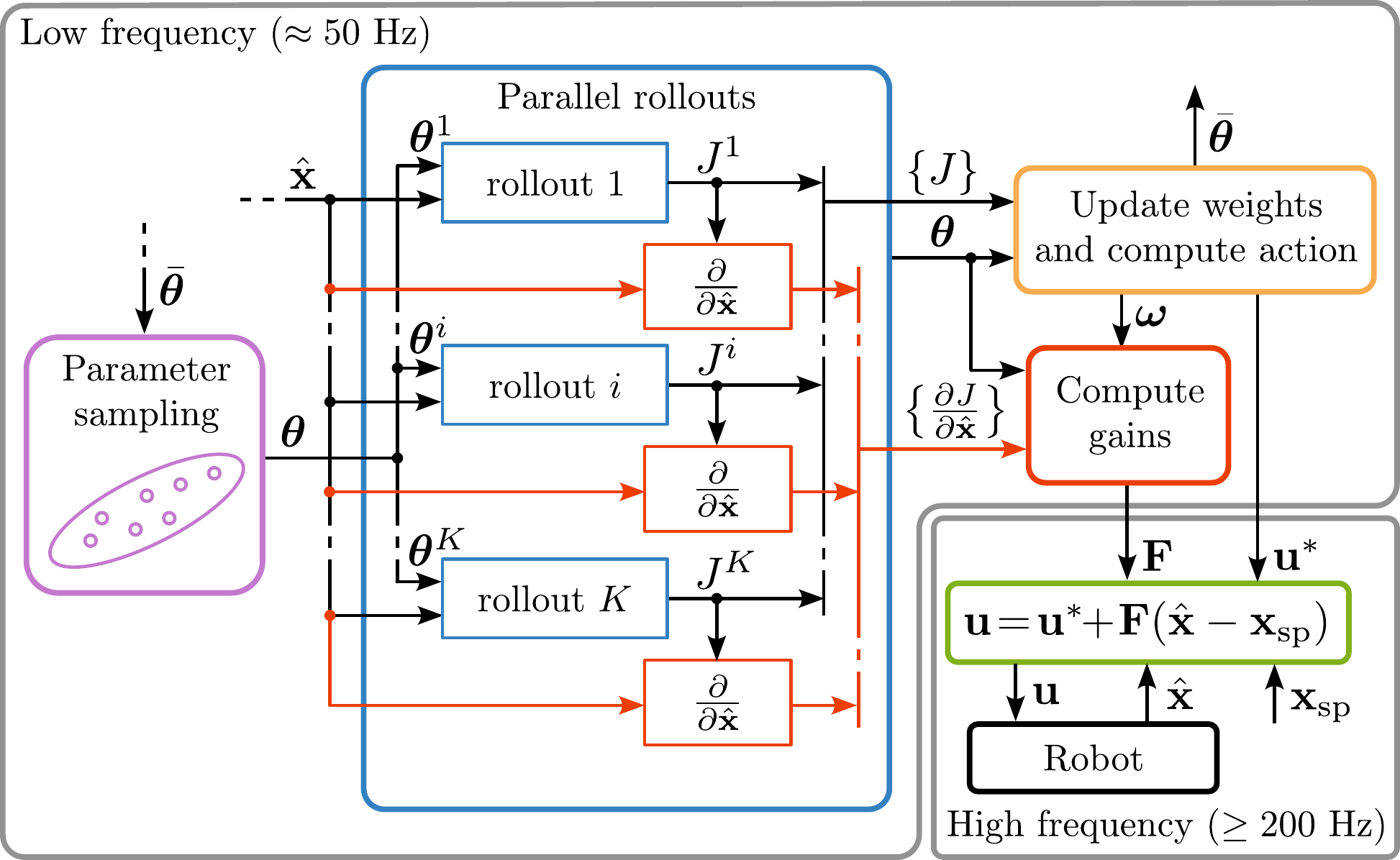}
    \caption{Block scheme describing the Feedback-MPPI method. 
    The proposed gain computation (red path) can leverage the same parallelization capabilities of the standard MPPI algorithm.
    Braces represent the stack of multiple signals.}
    \label{fig:block_scheme}
\end{figure}

Gradient-based methods have achieved great efficiency by linearizing dynamics and constraints at each timestep, notably through Real-Time Iteration schemes~\cite{Diehl2005}.
To achieve higher control frequencies, some works proposed the use of equivalent feedback gains that locally approximate the MPC solution, enabling more efficient closed-loop control \cite{Dantec2022, Grandia_Farshidian_Ranftl_Hutter:2019, Li2025, Zavala_Biegler_2009}.
For instance, Dantec et al. \cite{Dantec2022} use sensitivity analysis in a purely model-based context to derive a high-frequency feedback controller for humanoid robots. Their method computes a first-order approximation of the MPC solution using Riccati gains extracted from a Differential Dynamic Programming (DDP) solver, enabling a fast inner-loop controller to operate at 2 kHz without re-solving the full MPC at each cycle. This effectively bridges the gap between slow optimal planning and fast low-level control.
Similar ideas have been exploited for quadruped locomotion \cite{Grandia_Farshidian_Ranftl_Hutter:2019, Li2025} and for MPC based on Non Linear Programming (NLP) \cite{Zavala_Biegler_2009}.
Building on this idea, Hose et al. \cite{Hose2024} extend sensitivity-based approximation to the learning domain. They propose a parameter-adaptive scheme in which a neural network is first trained to imitate the nominal MPC policy, and then locally corrected at runtime using first-order sensitivities of the MPC solution with respect to varying system parameters. This allows the learned controller to generalize across different system configurations without retraining, making it robust and suitable for deployment on embedded hardware with limited resources.
In a similar vein, Tokmak et al. \cite{Tokmak2024} introduce an automated method to approximate nonlinear MPC solutions while preserving closed-loop guarantees. Their approach, ALKIA-X, uses kernel interpolation augmented with local sensitivity information to construct an explicit, non-iterative approximation of the MPC control law.

An alternative to traditional gradient-based MPC methods is the Model Predictive Path Integral (MPPI) control, a stochastic, sampling-based technique capable of handling highly nonlinear, non-convex problems without explicit gradients. MPPI has been effectively applied to aggressive autonomous driving~\cite{Williams2016, Williams2017}, agile quadrotor maneuvers~\cite{Minarik2024}, and contact-rich quadruped locomotion~\cite{Carius2022, Turrisi2024}. However, MPPI's computational load and noisy actions remain challenges, and can cause oscillations during hovering \cite{Minarik2024}. Recent advancements such as GPU-accelerated computations~\cite{Turrisi2024}, low-pass filtering~\cite{Kicki2025}, and adaptive sampling using learned priors~\cite{howell2022} aim to enhance real-time applicability and stability.

To further address these challenges, this work proposes integrating local linear feedback approximations into sampling-based MPPI controllers, inspired by Riccati-based feedback strategies from gradient-based MPC. The proposed method (Fig.~\ref{fig:block_scheme}) enhances responsiveness without necessitating full re-optimization at each step, unlocking high-frequency feedback control.
The main contributions of this paper are:
\begin{itemize}
\item The introduction of a novel method for computing local linear feedback gains within MPPI, extending the Riccati-based feedback MPC framework to stochastic sampling-based controllers.
\item Demonstrating the effectiveness of the approach in approximating the MPPI action to provide high-frequency feedback, improving accuracy, and action smoothness.
\item 
The approach is validated on quadrupedal robots in simulation and on aerial drones in real-world experiments, both performing dynamic motions.
\end{itemize}
Our results bridge the gap between sampling-based and gradient-based MPC, demonstrating that local feedback-enhanced sampling-based MPC as a viable strategy for real-time, complex robotic systems.

The remainder of the paper is structured as follows: in Sect.~\ref{sec:methodology} we describe the proposed methodology, first presenting standard MPPI (Sect.~\ref{sec:methodology_background}) and then extending it by computing a local approximation of its optimal solution from local variations in the initial state (Sect.~\ref{sec:methodology_fmppi}). This is first validated by comparing it with a linear quadratic regulator for a toy problem (Sect.~\ref{sec:methodology_valid}) and, in Sect.~\ref{sec:Applications}, through simulations and experiments on two robotic platforms, a quadruped legged robot and a quadrotor aerial vehicle.

\section{Methodology}\label{sec:methodology}

\subsection{Background on Model Predictive Path Integral Control}\label{sec:methodology_background}

Model Predictive Path Integral (MPPI) control is a sampling-based stochastic optimization approach widely employed to solve optimal control problems characterized by nonlinear dynamics, complex, non-convex cost functions, and potentially non-differentiable constraints~\cite{Turrisi2024, Minarik2024}. MPPI addresses optimal control problems defined as follows
\begin{equation}\label{eq:ocp}
    \begin{aligned}
       \bftheta^\ast =  \underset{\bftheta}{\arg\min} & \quad \ell_N(\bfx_N) + \sum_{i=0}^{N-1} \ell_i(\bfx_i, \bfu_i) \\
        \text{s.t.} & \quad \bfx_0 = \hat\bfx \\
        & \quad \bfx_{i+1} = \bff(\bfx_i, \bfu_i) \\
        & \quad \bfu_i = \bfpi(\bftheta, \bfx_i, t_i),
    \end{aligned}
\end{equation}
where $\bftheta$ denotes the decision variables, or parameters, $\bfu_i$ the control inputs at time $t_i$, $\bfx_i$ represents the system state evolving according to the dynamics $\bff(\bfx, \bfu)$, $\ell_i$ denotes the cost functions defined over the control horizon $N$ and, finally, $\hat\bfx$ indicates the current feedback state.
For generality, we consider the control inputs as parametrized by a possibly state dependent function $\bfpi(\bftheta, \bfx, t)$ that maps sampled parameters $\bftheta$ into an input trajectory, possibly also clipping inputs to their upper/lower bounds. This can be used to decrease the dimensionality of the search space and to improve the smoothness of the input trajectory\cite{howell2022}. Possible choices are direct sampling (zero-order)~\cite{williams2018}, cubic splines~\cite{Turrisi2024}, or Halton splines \cite{Pezzato_2025}. 

In MPPI, parameter samples are drawn from a Gaussian distribution centered at an initial guess $\bar\bftheta$, which may be obtained 
 from the solution at the previous step, according to $\bftheta \sim \mathcal{N}(\bar\bftheta, \bfSigma)$.
Each parameter sample $\bftheta^k = \bar\bftheta + \Delta\bftheta^k$ generates a rollout by propagating the system dynamics forward in time, producing a set of candidate trajectories. Each trajectory is associated with a cost, with $J^{k}$ denoting the total cost associated with the $k$-th sampled trajectory:
\begin{equation}
J^{k} := \ell_N(\bfx_N^k) + \sum_{i=0}^{N-1} \ell_i(\bfx_i^k, \bfu_i^k).
\end{equation}
The optimal parameter update in MPPI is then computed using a weighted average of parameter perturbations:
\begin{equation}\label{eq:theta_variation}
    \bftheta^{\ast} = \bar\bftheta + \sum_{k=0}^{K} \omega^k\Delta\bftheta^k,
\end{equation}
with trajectory-specific weights $\omega^k$ determined by their costs $J^k$ as follows:
\begin{equation}\label{eq:weights}
\omega^k = \frac{\mu^k}{\sum_{j=1}^{K}\mu^j}, \text{ with } \mu^{k} = \exp{\left(-\frac{1}{\lambda} \left( J^{k} - \rho \right)\right)},
\end{equation}
where $\rho = \min_{k=1,\dots,K}J^{k}$, and $\lambda$ is a tuning parameter balancing exploration and exploitation~\cite{Minarik2024,rizzi_robust_2023}.
In practical implementations, MPPI utilizes GPU acceleration for parallelization of trajectory evaluations, significantly reducing computational time and facilitating real-time performance on complex systems such as agile aerial and legged robots~\cite{Minarik2024,Turrisi2024}.
The computed optimal parameters from \eqref{eq:theta_variation} directly yield the commanded control action via:
\begin{equation}\label{eq:input_parametrization}
\bfu^{\ast} = \bfpi(\bftheta^\ast, \hat\bfx, t_0).
\end{equation}
In standard MPPI frameworks, the resulting optimal action is commanded directly to the system and typically held constant over the sampling interval.

\subsection{Feedback MPPI}\label{sec:methodology_fmppi}
A notable drawback of MPC controllers, MPPI included, is that the control frequency is directly tied to the computational rate of the MPPI solution, which is constrained by the platform's processing capabilities. 
To address this, we introduce a method for computing a first-order approximation of the MPPI action. This approximation enables high-frequency feedback for systems demanding high control bandwidth, eliminating the need for an additional tracking controller.

In order to compute a local approximation of the MPPI action, we derive the sensitivity of the optimal inputs $\bfu^{\ast}$ in \eqref{eq:input_parametrization} to variations in the current initial state $\hat\bfx$. This quantity, denoted as $\bfF\in\RealSet^{n_u \times n_x}$, maps state variations into inputs and can be used as a local feedback control gain to obtain a new action of the kind
\begin{equation}
    \bfu = \bfu^\ast + \bfF\left( \hat\bfx - \bfx_{\rm sp} \right),
\end{equation}
where $\bfx_{\rm sp}$ represents a local setpoint acquired from the MPPI solution\footnote{In contrast to several works that simply set this setpoint as the initial state $\bfx_0$ at which the solution was computed, we draw inspiration from  \cite{Subburaman2024} and continuously update $\bfx_{\rm sp}$ by linearly interpolating between $\bfx_0$ and $\bfx_1$.} which will be tracked by the inner loop.

To derive an expression for the MPPI gain matrix $\bfF$, it is sufficient to differentiate the optimal solution obtained through importance sampling \eqref{eq:theta_variation}--\eqref{eq:weights}.
In fact, applying the chain rule to \eqref{eq:input_parametrization} leads to:
\begin{equation}\label{eq:gains_def}
    \bfF = \pdv{\bfu^{\ast}}{\bfx_0} = \pdv{\bfpi}{\bftheta}\pdv{\bftheta^\ast}{\bfx_0} + \pdv{\bfpi}{\bfx_0},
\end{equation}
where $\pdv{\bfpi}{\bftheta}$ and $\pdv{\bfpi}{\bfx_0}$ depend on the specific policy parametrization choice, e.g., direct sampling or cubic splines. 
Then, one has to differentiate the optimal parameters $\bftheta^{\ast}$.
Recalling \eqref{eq:theta_variation}:
\begin{equation}\label{eq:dtheta_dx}
    \pdv{\bftheta^{\ast}}{\bfx_0} = \sum_{k=0}^{K} \Delta\bftheta^k  \pdv{\omega^k}{\bfx_0}.
\end{equation}
Then, noting that
\begin{equation}
    \pdv{\mu^k}{\bfx_0} =  -\frac{\mu^k}{\lambda}\left( \pdv{J^k}{\bfx_0} -  \pdv{\rho}{\bfx_0}\right),
\end{equation}
applying the quotient rule to \eqref{eq:weights} and simplifying, yields
\begin{equation}
    \pdv{\omega^k}{\bfx_0} = \frac{\omega^k}{\lambda}\left(\pdv{J^k}{\bfx_0} - \sum_{j=1}^{K} \omega^j\pdv{J^j}{\bfx_0} \right).
\end{equation}
Plugging back to \eqref{eq:dtheta_dx} one has
\begin{equation}
    \pdv{\bftheta^{\ast}}{\bfx_0} = \sum_{k=0}^{K} \Delta\bftheta^k \frac{\omega^k}{\lambda}\left(\pdv{J^k}{\bfx_0} - \sum_{j=1}^{K} \omega^j\pdv{J^j}{\bfx_0} \right),
\end{equation}
which can be combined with \eqref{eq:gains_def} to obtain the desired MPPI gains, depending on the chosen control parametrization:
\begin{equation}\label{eq:gains_final}
    \bfF =  \sum_{k=0}^{K} \pdv{\bfpi}{\bftheta}\Delta\bftheta^k \frac{\omega^k}{\lambda}\left(\pdv{J^k}{\bfx_0} - \sum_{j=1}^{K} \omega^j\pdv{J^j}{\bfx_0} \right)   + \pdv{\bfpi}{\bfx_0}.
\end{equation}
For instance, in the case of direct control sampling, where $\bftheta = (\bfu_0, \dots, \bfu_{N-1})$, with $\pdv{\bfpi}{\bftheta} = (\bfI, \bfzero, \dots, \bfzero)$ and $\pdv{\bfpi}{\bfx_0} = \bfzero$:
\begin{equation}\label{eq:gains_zero-order}
    \bfF = \sum_{k=0}^{K} \Delta\bfu^k_0 \frac{\omega^k}{\lambda}\left(\pdv{J^k}{\bfx_0} - \sum_{j=1}^{K} \omega^j\pdv{J^j}{\bfx_0} \right).
\end{equation}
Similar expressions can be derived also in the case of different input parametrization.

\begin{remark}\label{rem:1}
One of the advantages of sampling-based MPC over gradient-based MPC is the ability to optimize over discontinuous optimal control problems. Since our gain computation relies on the gradients of the rollout costs, we do require local differentiability of the dynamics and costs in \eqref{eq:ocp}. This is, however, not a limitation: in fact, most robotic applications are described by at least piecewise smooth functions. Indeed, such gains will only approximate the local behavior of the system away from discontinuities. 
This is similar to the fact that linear MPC approximations do not track active set changes when dealing with inequality constraints \cite{Zavala_Biegler_2009}. Still, while computing gains requires local differentiability of the rollouts, the ability of the solver to explore the solution space and optimize over discontinuous non-convex problems is not affected.
\end{remark}

\begin{remark}\label{rem:gains_constraints}
In many MPPI implementations, including ours, generic inequality constraints can be managed using indicator functions by adding a large constant cost to trajectories that violate the constraint. Since this cost term is (locally) constant, it does not influence the gain computation, meaning the feedback system remains unaware of such constraints. A straightforward way to overcome this limitation is to incorporate constraints through smooth barrier functions, as demonstrated in Feedback-MPC \cite{Grandia_Farshidian_Ranftl_Hutter:2019}. 
On the other hand, since input constraints are treated by directly clipping the sampled trajectory inside the rollouts, MPPI gains will naturally account for such constraints if active.
\end{remark}

\subsection{Implementation}

Notably, the proposed gain computation \eqref{eq:gains_final} still makes use of parallel computation to achieve real-time performance even with a large number of samples. In fact, the cost gradients $\pdv{J^k}{\bfx_0}$ can be evaluated separately for each rollout and later combined with the appropriate trajectory weights $\omega^k$.
The whole Feedback-MPPI algorithm has been implemented in a standalone Python library that we released as open source. This library makes use of JAX to run JIT compiled code on CUDA-compatible GPUs to perform parallel rollouts. Moreover, we make use of the Automatic Differentiation capabilities provided by JAX to compute the cost gradients in \eqref{eq:gains_final} in parallel together with the rollouts, improving the computational efficiency of the method.

\subsection{Illustrative example}\label{sec:methodology_valid}

To assess the efficacy of the proposed method in determining the optimal gains for the MPPI solution, we first illustrate its behavior through a quantitative example involving an inverted pendulum system. These gains are compared to those obtained by applying an infinite horizon discrete-time Linear Quadratic Regulator (LQR) to the linearized system around the upward equilibrium. Although a formal proof of the relationship between MPPI and Riccati gains is not available at the time of writing, it is reasonable to assume that the methods should yield similar gains under similar circumstances.

\begin{figure}[t!]
    \centering
\includegraphics[width=0.98\columnwidth]{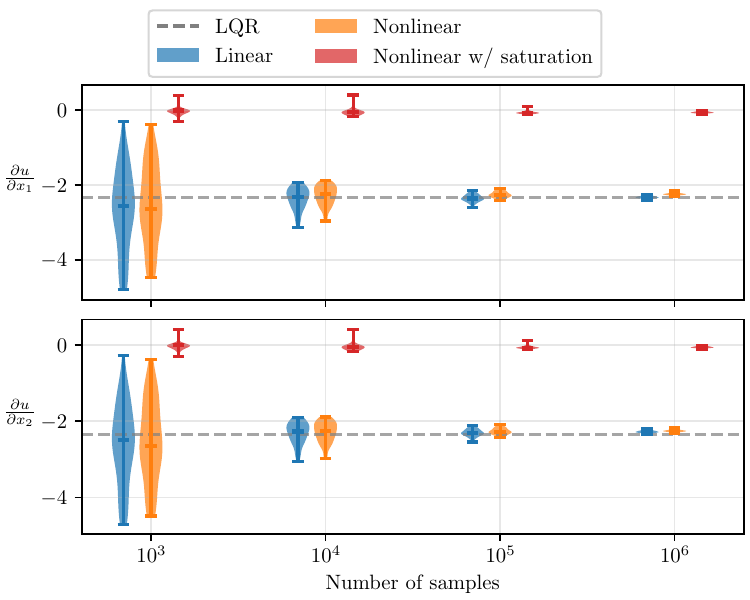}
    \caption{Comparison between MPPI gains obtained for the inverted pendulum system under different conditions (linear/nonlinear rollout, input saturation) for an increasing number of samples, and LQR gains computed around the upward equilibrium (dashed gray line).}
    \label{fig:lqr_gains_comparison}
\end{figure}

In this scenario, the algebraic Riccati equation was solved using $\bfQ = \bfI$ and $\bfR = 1$, resulting in the optimal Riccati gain matrix $\bfF_{\rm LQR}$ and the solution matrix $\bfS$.
Next, we formulated an equivalent MPPI problem, aligning the running cost with that of the LQR and setting the terminal cost as $\ell_N = \bfx^T \bfS \bfx$ to solve the infinite horizon problem. Three cases are considered for MPPI: first, using the same linearized dynamics as the LQR, then with the nonlinear inverted pendulum dynamics and, finally, including input bounds to saturate the optimal solution.
The MPPI was then solved multiple times from an initial condition $\bfx_0 = (\pi/8, 0.0)$, with gains computed using \eqref{eq:gains_final}. This procedure was repeated while varying the number of samples $K$.
The results in Fig.~\ref{fig:lqr_gains_comparison} demonstrate that in the unconstrained cases (Linear and Nonlinear) MPPI gains $\bfF = \left(\pdv{u}{x_1}, \pdv{u}{x_2}\right)$ are on average similar to LQR optimal gains for both linear and nonlinear rollouts, with a variance shrinking significantly as the number of samples increases. As the gain computation relies on the linearization around the rollout trajectories, minor differences can be ascribed to the different state trajectories obtained with the linearized model.
In the input constrained case, MPPI gains instead converge to zero: this effect, which correctly predicts the fact that no additional feedback action can be added to the control, stems from the fact that the partial derivative $\pdv{\bfpi}{\bftheta}$ collapses to zero when inputs are clipped.

\section{Applications}\label{sec:Applications}

Riccati-like gains have consistently demonstrated their effectiveness in various applications through numerous studies and practical implementations \cite{Dantec2022, Grandia_Farshidian_Ranftl_Hutter:2019, Li2025, Subburaman2024, Belvedere2025}. Given this well-established track record, our objective in this section is not to revalidate their utility. Instead, we aim to illustrate how these gains can be successfully integrated within the framework of sampling-based MPC running at low frequency to dynamically control two robotic platforms, and how the additional computational effort balances with performance gains even when compared to MPPI running at higher frequencies.

\subsection{Quadruped motion control}

We first test the proposed method on a simulated quadruped robot based on Aliengo\footnote{Aliengo quadruped robot: https://www.unitree.com/products/aliengo}, a 24kg electric quadruped developed by Unitree.
The simulations are performed in MuJoCo, running on a laptop with an Intel i7-13700H CPU and an Nvidia 4050 6GB GPU. All relevant settings are reported in Tab.~\ref{tab:MPPI_settings_quadruped}.

The dynamical model of the system adopts the simplified Single Rigid Body Dynamics model formulation \cite{Turrisi2024}. This model focuses on the quadruped's core translational and rotational movements, omitting the swinging legs' dynamics given their typically small mass in comparison with the trunk of the robot. The robot's dynamics is centered around the CoM frame, and is described using two reference frames: an inertial frame $\mathcal{W}$, and a body-aligned frame $\mathcal{C}$ at the Center of Mass (CoM). The state vector is
\[
\bfx = \bigl(\bfr_\mathrm{c},\,\bfv_\mathrm{c},\,\bfphi,\,\bfomega\bigr)
      \in \RealSet^{3}\times \RealSet^{3}\times \RealSet^{3}\times \RealSet^{3},
\]
with position $\bfr_\mathrm{c}=(x,y,z)$ and velocity $\bfv=(v_x,v_y,v_z)$ expressed in the inertial frame,  
the robot body orientation $\bfphi = (\phi, \theta, \psi)$ where $\phi, \theta, \psi,$ are the roll, pitch, and yaw respectively, and the angular velocity $\bfomega=(\omega_x,\omega_y,\omega_z)$ expressed in body frame. Finally, the equations of motion read as
\[
\begin{aligned}
&\dot{\bfr}_{\mathrm{c}}     = \bfv_{\mathrm{c}}\\
&\dot\bfv_{\mathrm{c}}     = \dfrac{1}{m}\, \sum_{i=1}^4 \delta_\mathrm{i} \bff_\mathrm{i} + \bfg\\
&\dot{\bfphi}     = \bfE^{\prime-1}(\mathcal{\bfphi}) \bfomega \\
&\dot{\bfomega} = - \bfI_{\mathrm{c}}^{-1}\left( \bfomega \times{ } \bfI_{\mathrm{c}}\right)  \bfomega +\sum_{i=1}^4 \delta_\mathrm{i} \bfI_{\mathrm{c}}^{-1}{ } \bfp_\mathrm{i} \times{ }\bff_\mathrm{i},
\end{aligned}
\]
with $m$ the robot mass, $\bfg$ the gravitational acceleration and $\bfI_c \in \mathbb{R}^{3 \times 3}$ the constant inertia tensor centered at the robot's CoM; $\bfE^{\prime-1}$ is a mapping from the robot's angular velocity to Euler rates; $\bfp_\mathrm{i} \in \mathbb{R}^3$ is the displacement vector between the CoM position $\bfr_\mathrm{c}$ and the $i$-th robot's foot. Binary variables $\delta_i = \{0, 1\}$, extracted from a precomputed periodic gait sequence, indicate whether an end-effector makes contact with the environment and can produce Ground Reaction Forces $\bff_\mathrm{i}$ (GRF), which we optimize in our controller formulation.

\begin{table}[t]
\renewcommand{\arraystretch}{1.0}
\centering
\caption{MPPI settings for the Quadruped Experiments}
\begin{tabular}{c|c|cc}
    \hline
    & \textbf{Symbol} & \multicolumn{2}{c}{\textbf{Value}} \\
    \hline
    \multirow{5}{4em}{\centering Solver} & $N$ & \multicolumn{2}{c}{$10$~steps} \\
    & $\delta_t$ & \multicolumn{2}{c}{$0.02$~s} \\
    & $K$ & \multicolumn{2}{c}{$5000$~samples} \\
    & $\bfSigma$ & \multicolumn{2}{c}{$9\bfI$} \\
    & $\lambda$ & \multicolumn{2}{c}{1} \\
    \hline
    \multirow{7}{4em}{\centering Weights} & & \textbf{Running} & \textbf{Terminal} \\
     & $\bfQ_{\rm r}$ & $\diag(0, 0, 1500)$ & $\diag(0, 0, 1500)$  \\
    & $\bfQ_{\rm v}$ & $\diag(200, 200, 200)$ & $\diag(200, 200, 200)$  \\
    & $\bfQ_{\rm \phi}$ & $\diag(1500, 1500, 0)$ & $\diag(1500, 1500, 0)$ \\
    & $\bfQ_{\omega}$ & $\diag(20, 20, 50)$ & $\diag(20, 20, 50)$ \\
    & $\bfR$ & $10^{-4}\bfI$ & --- \\
    \hline
\end{tabular}\label{tab:MPPI_settings_quadruped}
\end{table}

\subsubsection{Simulation results} In the simulation, we ask the robot to track a set of linear velocity references (between $0$,  $0.5$ m/s) over a randomly generated rough terrain while being subject to random disturbances (between $\pm 5$ Nm). For this task, the cost function is designed to track a desired velocity reference (linear and angular) while maintaining a desired posture (height, roll, and pitch of the robot).
This is achieved by incorporating a weighted quadratic state error plus a regularization term for the GRF (gravity compensation)
\begin{equation}
    \ell_{\rm track} = \|\bfx - \bfx_{\rm track}\|^2_{\bfQ} + \|\bfu - \bfu_{\rm reg}\|^2_{\bfR}.
\end{equation}
The control inputs (GRFs) are sampled using linear splines and are clipped to enforce friction cone constraints for non-slipping conditions \cite{Turrisi2024}.

\begin{figure}[!t]
    \includegraphics[width=0.9\columnwidth]{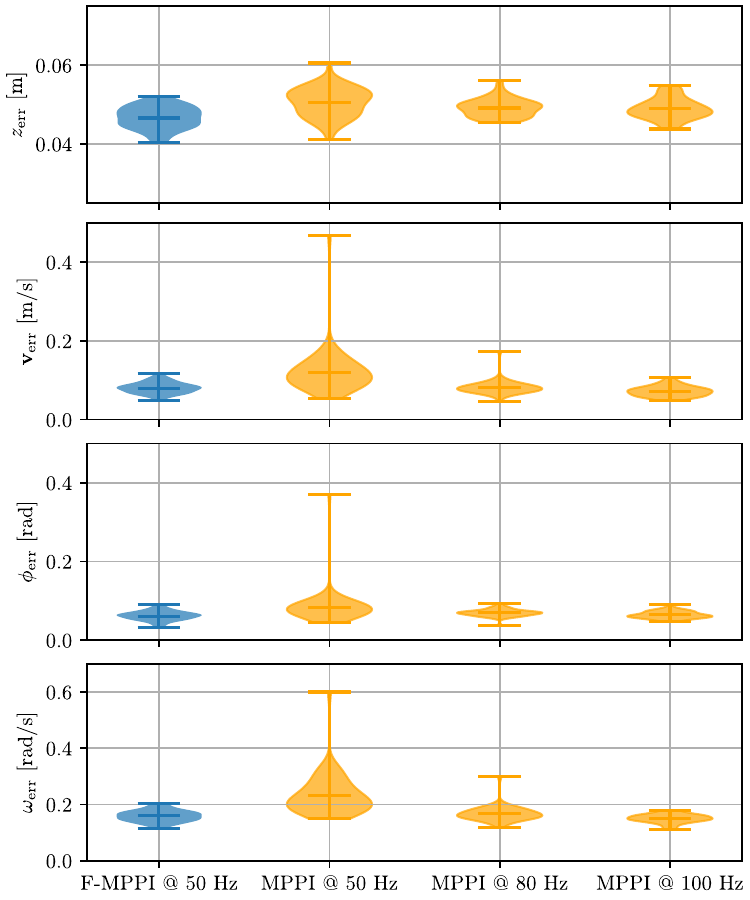}
    
    \hspace{0.5cm}
    \includegraphics[width=0.85\columnwidth]{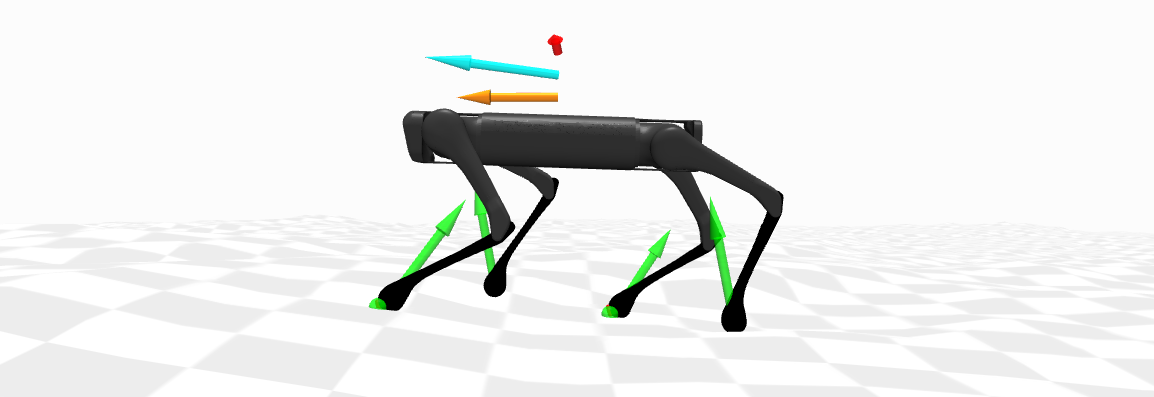}
    
    \caption{Top: Tracking results for the quadruped simulation of F-MPPI against MPPI running at three different control frequencies; Bottom: A snapshot of the quadruped robot tracking a desired velocity (orange arrow) on a randomly generated uneven terrain, while being perturbed by an external disturbance (red arrow). Green arrows represent the GRFs, while the blue arrow depicts the actual robot's velocity.}
\label{fig:results_quadruped}
\end{figure}
We run F-MPPI at 50Hz, and compare its performance to standard MPPI at three different control frequencies, 50Hz, 80Hz, and 100Hz. We selected 50Hz as base frequency to be consistent with the real quadrotor experiments in Sect.\ref{sec:quadrotor_exp}, where such frequency is limited by the onboard hardware, and to simulate the use of a similar, less powerful, embedded system. Then, we progressively increased the MPPI frequency to reflect the fact that the increased computational cost of F-MPPI relatively decreases the maximum control frequency for a given computational budget. In all cases, we sample 5000 parameters, bringing the computational complexity of F-MPPI to $3$~ms per run, while that of MPPI to $2$~ms. Our goal is to show that utilizing the MPPI gains brings benefits to the final performance of the system, even when accounting for a decrease in the control frequency due to the additional computations.  In all cases, a low-level controller that runs at 500Hz maps the GRFs to joint torques via a simple Jacobian mapping (see \cite{Turrisi2024}). When F-MPPI is used, this mapping consistently relies on GRFs updated at the same rate (500Hz), leveraging the gains calculated at a lower frequency by our algorithm. 

The final results, which report the Mean Absolute Errors over 50 trials, are shown in Fig.~\ref{fig:results_quadruped}. F-MPPI, computed at 50 Hz, remains competitive even against an MPPI running two times faster ($100$~Hz), and outperforms MPPI running at a similar computational budget ($80$~Hz).

\subsection{Quadrotor motion control}

The second robotic platform consists of a quadrotor robot based on MikroKopter and running on the Telekyb3 framework for the hardware interface and state estimation. 
An Unscented Kalman Filter implemented in Telekyb3 is used to fuse the IMU measurements and the motion capture position feedback to obtain an estimate of the state $\hat{\bfx}$.
First, the proposed method has been tested in a Software-In-The-Loop simulation using a Gazebo environment where the robot firmware, state estimation, and motor dynamics are simulated to mimic real-world experiments.
The simulation runs on a laptop with an Intel i7-13700H CPU and an Nvidia A1000 6GB GPU.
In the experiments, the F-MPPI controller is fully executed \emph{onboard} on an Nvidia Jetson Orin NX 16GB. 
The same implementation and MPPI settings used in the simulation have been employed in the experiments, with only minor differences in the hardware interface. All relevant settings are reported in Tab.~\ref{tab:MPPI_settings}.
\begin{table}[t]
\renewcommand{\arraystretch}{1.0}
\centering
\caption{MPPI settings for the Quadrotor Experiments}
\begin{tabular}{c|c|cc}
    \hline
    & \textbf{Symbol} & \multicolumn{2}{c}{\textbf{Value}} \\
    \hline
    \multirow{5}{4em}{\centering Solver} & $N$ & \multicolumn{2}{c}{$15$~steps} \\
    & $\delta_t$ & \multicolumn{2}{c}{$0.05$~s} \\
    & $K$ & \multicolumn{2}{c}{$800$~samples} \\
    & $\bfSigma$ & \multicolumn{2}{c}{$25\bfI$} \\
    & $\lambda$ & \multicolumn{2}{c}{1} \\
    \hline
    \multirow{7}{4em}{\centering Weights} & & \textbf{Running} & \textbf{Terminal} \\
     & $\bfQ_{\rm r}$ & $\diag(100, 100, 125)$ & $\diag(100, 100, 125)$  \\
    & $\bfQ_{\rm v}$ & $\diag(0.5, 0.5, 2.5)$ & $\diag(10, 10, 25)$  \\
    & $\bfQ_{\rm q}$ & $\diag(0.5, 0.5, 50)$ & $\diag(10, 10, 100)$ \\
    & $\bfQ_{\omega}$ & $\diag(0.25, 0.25, 25)$ & $\diag(0.5, 0.5, 50)$ \\
    & $\bfR$ & $10^{-2}\bfI$ & --- \\
    & $Q_{\rm obs}$ & $10^6$ & $10^6$ \\
    \hline
\end{tabular}\label{tab:MPPI_settings}
\end{table}

\begin{figure*}[!t]
    \centering
    \includegraphics[width=0.975\linewidth]{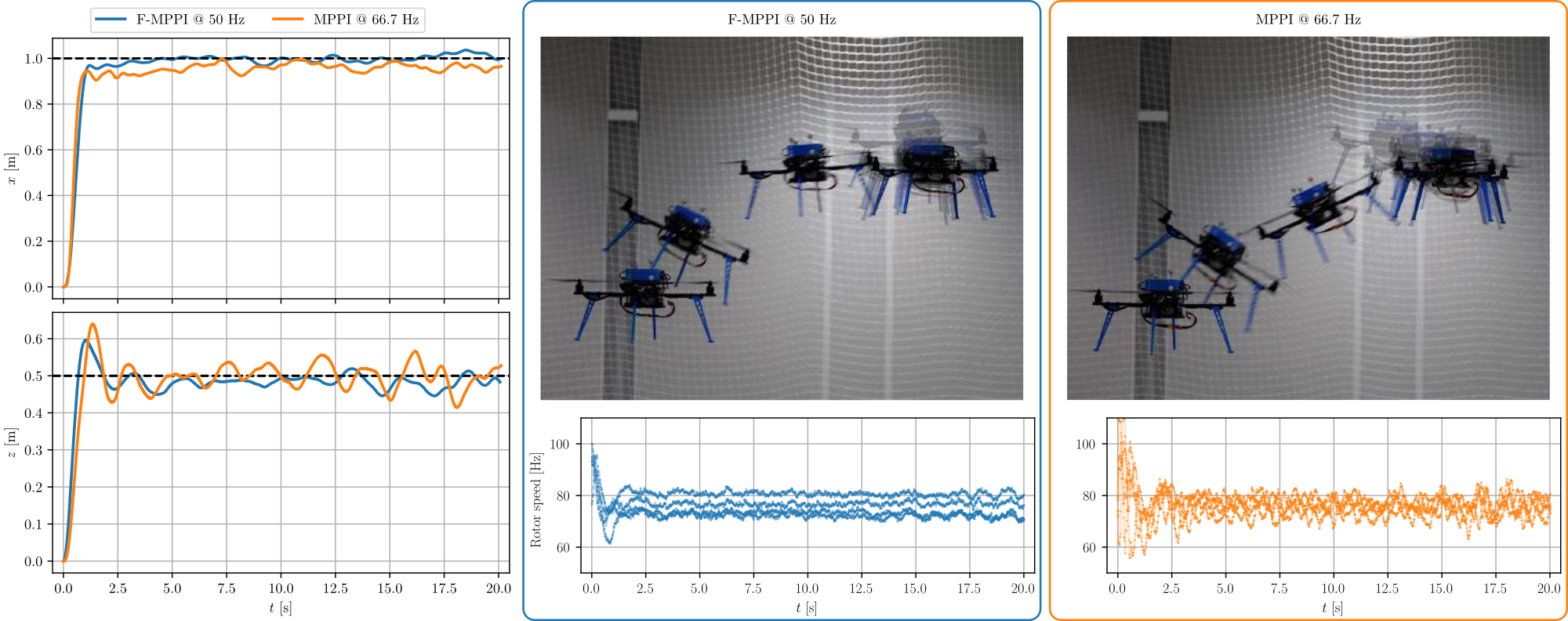}
    \caption{Evolution of the quadrotor position in the $x\!\!-\!\!z$ plane, stroboscopic view of the quadrotor motion and commanded rotor speeds during the experiment. In blue is the proposed framework, while in orange is the standard MPPI controller running at the maximum frequency allowed by the hardware platform.
    Thanks to the proposed method, the resulting control action is smoother and the motion more precise, with a reduction of $64.6$\% and $28.9$\% in RMSE for the $x$ and $z$ components, respectively.
    See the supplementary material for a clip of the experiment.}
    \label{fig:panel}
\end{figure*}

Following the scheme depicted in Fig.~\ref{fig:block_scheme}, the F-MPPI algorithm operates at a frequency of $50$~Hz, while the low-level control module, based on the MPPI gain matrix, updates the control action at a frequency of $200$~Hz.

We adopt the standard rigid‑body model of a quadrotor whose center of mass coincides with its geometric center\cite{AmCaArRoChFr:18}.    
The state vector is
\[
\bfx = \bigl(\bfr,\,\bfv,\,\bfq,\,\bfomega\bigr)
      \in \RealSet^{3}\times \RealSet^{3}\times \mathbb S^{3}\times \RealSet^{3},
\]
with position $\bfr=(x,y,z)$ and velocity $\bfv=(v_x,v_y,v_z)$ expressed in the inertial frame,  
the unit quaternion $\bfq=(q_w,q_x,q_y,q_z)$ describing the body attitude and the angular velocity $\bfomega=(\omega_x,\omega_y,\omega_z)$ expressed in body frame, and $\mathbb S^{3}$ representing the unit 3-sphere.

With the rotor speeds as inputs
\(
\bfu=(w_1,w_2,w_3,w_4),
\)
the total thrust $f$ and control torque $\bftau$ generated in body frame are obtained through the allocation matrix $ (f,\,\bftau) = \bfT(l,k_f,k_m) \bfu^2$,
where $l$ denotes the quadrotor arm length and $k_f$, $k_m$ the propeller aerodynamic coefficients \cite{AmCaArRoChFr:18}.
The equations of motion therefore read
\[
\begin{aligned}
\dot{\bfr}     &= \bfv, & \dot{\bfq} &= \dfrac{1}{2}\,\bfq\otimes\begin{bmatrix}0\\\bfomega\end{bmatrix},\\
\dot{\bfv}     &= \dfrac{1}{m}\,\bfR(\bfq)\begin{bmatrix}0\\ 0 \\ f\end{bmatrix} + \bfg, & \quad
\dot{\bfomega} &= \bfI_c^{-1}\bigl(\bftau - \bfomega \times \bfI_c\,\bfomega\bigr),
\end{aligned}
\]
where $\bfR(\bfq)\in\mathrm{SO}(3)$ is the rotation matrix associated with $\bfq$,  
$m$ is the total mass, $\bfg$ the gravity force vector, and
\(
\bfI_c
\)
the inertia tensor centered at the robot CoM.

Both the simulation and experiments use the same base cost function designed to reach a position goal while trading off control effort.
The cost function term for achieving the goal incorporates a weighted quadratic state error:
\begin{equation}
    \ell_{\rm goal} = \|\bfr - \bfr_{\rm goal}\|^2_{\bfQ_r} + \|\bfv\|^2_{\bfQ_v} + \|\bfq - \bfq_{0}\|^2_{\bfQ_q} + \|\bfomega\|^2_{\bfQ_\omega},
\end{equation}
where $\bfq_{0} = (1,0,0,0)$.
Inputs are sampled using cubic splines with $N_{\rm knot} = 5$ knot points. The sampled trajectory is then clipped to enforce input limits.
To minimize control effort and regularize the solution, the input is kept close to the constant hovering input $\bfu_h$ with a cost:
\begin{equation}
    \ell_{\bfu} = \|\bfu - \bfu_{h}\|^2_{\bfR}.
\end{equation}

\subsubsection{Experimental results}\label{sec:quadrotor_exp}

The effectiveness of the F-MPPI method is now demonstrated in a real-world scenario where computational resources are also limited.
Specifically, we execute standard MPPI on the onboard computer at its maximum feasible frequency of 66.7 Hz. During these tests, the quadrotor is tasked with reaching a goal displaced of $(1, 0.5)$~m in the $x$-$z$ plane. The overall system is tuned for best performance with the standard MPPI running at 66.7 Hz, while the frequency is lowered to 50 Hz for F-MPPI to meet real-time constraints without modifying the problem.

The experimental results, illustrated in Fig.~\ref{fig:panel}, highlight that the proposed F-MPPI method not only reaches the target more quickly but also with superior precision compared to standard MPPI. To quantitatively evaluate performance, we calculate the Root Mean Square Error (RMSE) for both the $x$ and $z$ components after the initial transient phase. The F-MPPI method achieves RMSE values of $0.017$~m and $0.027$~m for the $x$ and $z$ components, respectively, while the standard MPPI yields higher RMSEs of $0.048$~m and $0.038$~m.

An additional advantage of F-MPPI is the significantly smoother control commands it generates, as shown in Fig.~\ref{fig:panel} (bottom right). This smoothness can have practical benefits, such as reducing actuator wear and lowering power consumption, thus enhancing the overall efficiency and longevity of the system.

\subsubsection{Computation time analysis}

\begin{figure}[!t]
    \centering
    \includegraphics[width=0.9\columnwidth]{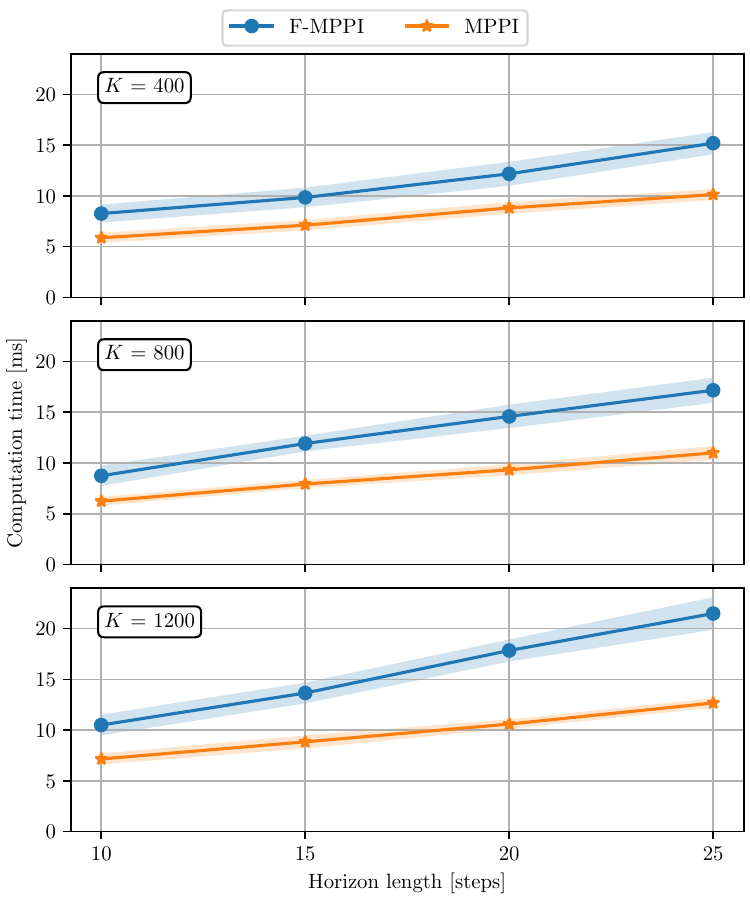}
    \caption{Computation times for the proposed method (F-MPPI) and standard MPPI in the quadrotor regulation problem, analyzed across varying control horizons and numbers of parallel samples. The analysis was conducted on an Nvidia Jetson Orin NX 16GB with 1024 GPU cores.}
    \label{fig:timings_quadrotor}
\end{figure}

In order to study the impact of the proposed method on the computation time, we have collected data by running several simulations of the quadrotor on the onboard Jetson computer with and without the MPPI gains calculation. Results obtained with different control horizons $N$ and number of samples $K$ are reported in Fig.~\ref{fig:timings_quadrotor}. 
The difference between F-MPPI and standard MPPI can be ascribed to the gain computation, which increases overall runtime by roughly $40\%$--$70\%$.
The overhead grows with the number of samples and the prediction horizon; it reaches the high end of this range when 
$K=1200$, where the rollouts outnumber the available GPU cores.

It should be recognized that this approach has a higher computational burden compared to methodologies developed for gradient-based MPC that rely on Riccati recursion \cite{Dantec2022} or NLP sensitivities \cite{Belvedere2025}.
Ultimately, the performance gains will be determined by the specific system characteristics that dictate whether the additional computational cost is worthwhile.
Nevertheless, we demonstrated that the proposed method is a viable option for sampling-based MPC. As evidenced by the experimental results, even running F-MPPI at the low frequency of $50$~Hz can yield significant performance improvements compared to standard MPPI thanks to the high-frequency feedback provided by the MPPI gains implemented in the inner-loop.

\subsection{Application to hybrid systems and non-convex constraints}

As the proposed method relies on differentiating the rollout trajectories, MPPI gains only provide a first-order approximation of the optimal solution (cfr. Remark~\ref{rem:1}). However, we demonstrate that this limitation does not hinder the method’s applicability to discontinuous problems arising from hybrid dynamical systems or the presence of inequality constraints. 

\begin{figure}[t]
    \centering
    \vspace{5mm}
    \hspace{-5mm}
    \includegraphics[width=0.9\columnwidth]{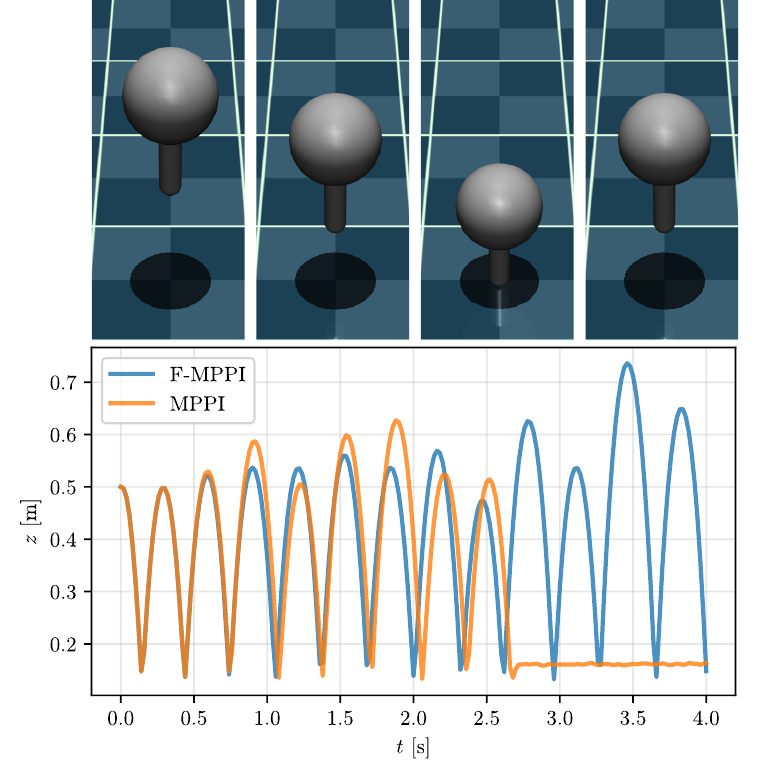}
    \vspace{-3mm}
    \caption{Comparison of the height evolutions of the 1D hopper robot under the two different control laws. The results were obtained by directly performing sampling and differentiation through the dynamics simulator.}
    \label{fig:hopper}
\end{figure}

\textbf{1D Hopper.}
The method was applied to a 1D Hopper robot whose hybrid dynamics is simulated through Mujoco MJX. This robot can extend its leg to push on the ground and propel upwards. However, due to the leg extension being limited, it has to resort to a bouncing motion to reach higher positions.
In this setting, we have implemented the F-MPPI method running at $50$~Hz with gains closing the low-level loop at $500$~Hz. 
The robot starts from the desired height of $0.5$~m and repeatedly bounces around the same height. To mimic a real-world scenario, the velocity of the robot is perturbed with Gaussian noise. Figure~\ref{fig:hopper} shows results with and without the use of MPPI gains. Without the additional correction provided by the inner loop, disturbances make the robot enter a local minimum, remaining stuck to the ground. On the other hand, the proposed method is able to complete the motion, demonstrating how MPPI gains are able to locally regulate the system across the two hybrid modes.

\begin{figure}[t]
    \centering
    \hspace{-5mm}
    \includegraphics[width=0.9\columnwidth]{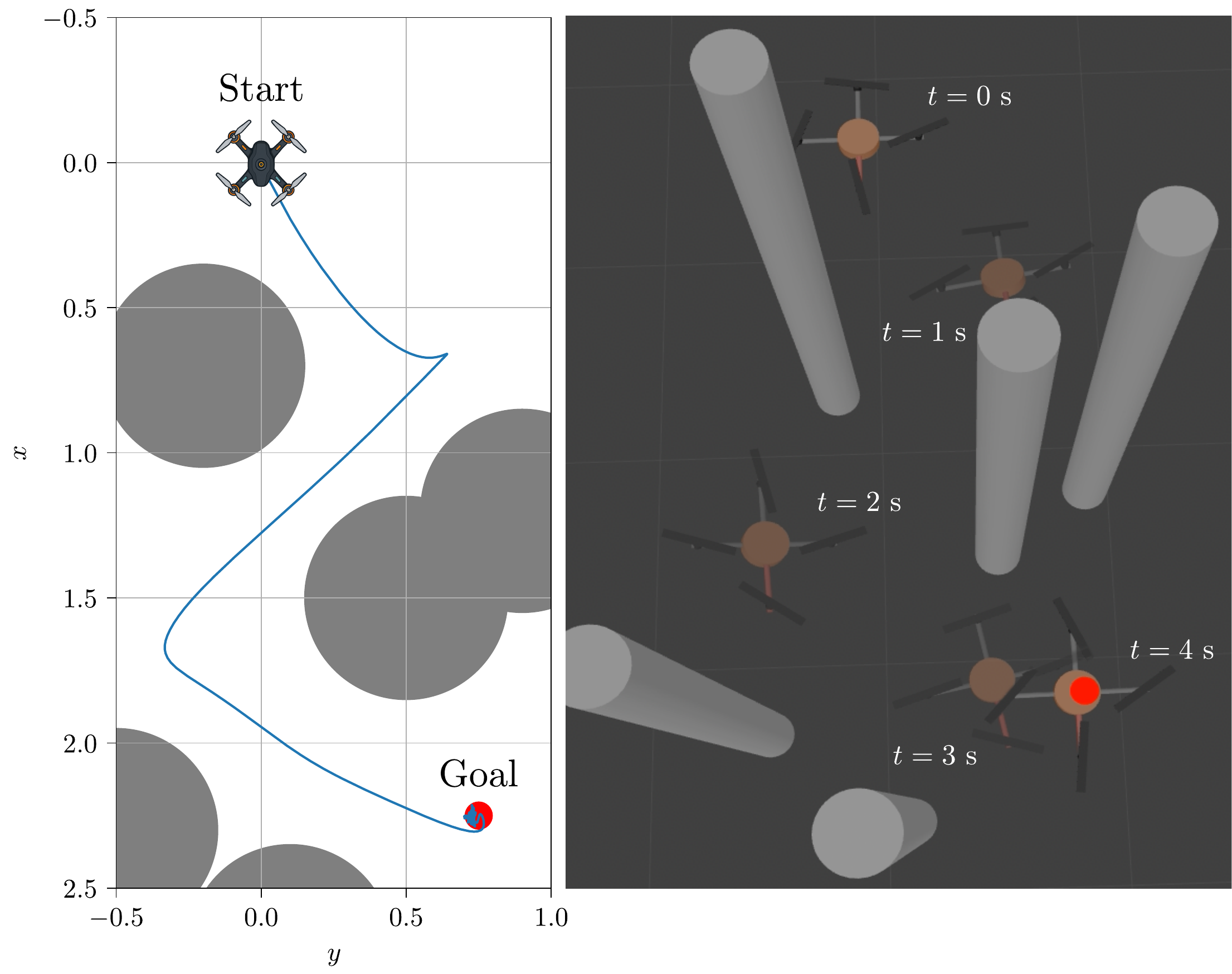}
    \caption{Path executed by the robot when navigating among obstacles. Top-down view with obstacles' radii augmented by the quadrotor dimension (left). Stroboscopic view of the motion in the Gazebo simulator (right). See the supplementary material for a clip of the simulation.}
    \label{fig:navigation}
\end{figure}

\textbf{Quadrotor navigation among obstacles.}
Despite generic constraints not being explicitly accounted for in the gain computation (cfr. Remark \ref{rem:gains_constraints}), the method can effortlessly incorporate them and handle non-convex optimization problems.
To illustrate this, we conducted simulations involving navigating through a series of stationary circular obstacles to reach the designated goal position.
Following \cite{Minarik2024}, the obstacle avoidance constraint is enforced by incorporating a barrier term into the cost function. This significantly penalizes trajectories that collide with the environment. The barrier term can be added to the cost as
\(
\ell_{\rm obs} = Q_{\rm obs}\,\bm{1}_{\bfx\in\chi_{\rm obs}},
\)
where $\chi_{\rm obs}$ represents the space occupied by obstacles.

Figure~\ref{fig:navigation} depicts the path executed in the navigation scenario. The robot successfully navigates to the goal in approximately $4$~s, avoiding the obstacles or getting stuck in local minima, before hovering at the goal position.

\section{Conclusion and Future Work}

In this letter, we introduced a way to efficiently compute an approximation of the sampling-based MPPI predictive control method in the form of linear feedback gains. This has allowed to extend the typical paradigm used in DDP-like and gradient-based methods for producing high-frequency MPC actions through the use of Riccati gains. The effectiveness of the method has been demonstrated on two robotic platforms, a legged robot and an aerial vehicle, through simulations and experiments. Results show that, similarly to related works, introducing such high-frequency MPPI approximation provides a benefit over plain MPPI, even when accounting for the higher computational cost. This development paves the way to several possible advancements in sampling-based MPC where such gains could be embedded, such as sensitivity-based robust MPC \cite{Belvedere2025}, safe reinforcement learning \cite{Zanon_Gros_2021}, and data augmentation for sample efficient learning \cite{Tagliabue_How_2024}.
Furthermore, future research will delve into computational aspects to streamline the gains calculation, for instance, optimizing the backpropagation process or further leveraging parallelization. Additionally, techniques to incorporate inequality constraints into the gains calculation will be investigated.

\section*{Acknowledgment}
The authors express their gratitude to Odichimnma Ezeji for her assistance in implementing obstacle avoidance in the quadrotor simulation.

\bibliographystyle{IEEEtran}
\bibliography{IEEEabrv, refs}

\end{document}

%% file: mathsym.tex
\def\RealSet{\mathbb{R}}

\def\diag{\mathrm{diag}}

\def\bfzero{{\mathbf{0}}}

\def\bff{{\mathbf{f}}}
\def\bfg{{\mathbf{g}}}

\def\bfp{{\mathbf{p}}}
\def\bfq{{\mathbf{q}}}
\def\bfr{{\mathbf{r}}}

\def\bfu{{\mathbf{u}}}
\def\bfv{{\mathbf{v}}}

\def\bfx{{\mathbf{x}}}

\def\bfE{{\mathbf{E}}}
\def\bfF{{\mathbf{F}}}

\def\bfI{{\mathbf{I}}}

\def\bfQ{{\mathbf{Q}}}
\def\bfR{{\mathbf{R}}}
\def\bfS{{\mathbf{S}}}
\def\bfT{{\mathbf{T}}}

\def\bfSigma{{\bm{\Sigma}}}

\def\bftheta{{\bm{\theta}}}

\def\bfpi{{\bm{\pi}}}

\def\bftau{{\bm{\tau}}}

\def\bfphi{{\bm{\phi}}}

\def\bfomega{{\bm{\omega}}}